\documentclass[letterpaper, 10 pt, conference]{ieeeconf}  

\IEEEoverridecommandlockouts                              

\usepackage{graphicx}
\usepackage{epsfig} 
\usepackage{mathptmx} 
\usepackage{times} 
\usepackage{amsmath} 
\usepackage{amssymb}  
\usepackage{float}  %
\usepackage{subfigure}  %
\usepackage{algorithm} 
\usepackage{algorithmicx} 
\usepackage[noend]{algpseudocode}
\usepackage{multirow} 
\usepackage{amsmath} 
\usepackage{xcolor}
\usepackage{epstopdf}
\usepackage{cite}
\usepackage{amsfonts}
\usepackage{textcomp}

\makeatletter
\newcommand*\bigcdot{\mathpalette\bigcdot@{.5}}
\newcommand*\bigcdot@[2]{\mathbin{\vcenter{\hbox{\scalebox{#2}{$\m@th#1\bullet$}}}}}
\makeatother

\title{\LARGE \bf
RF-LIO: Removal-First Tightly-coupled Lidar Inertial Odometry in High Dynamic Environments
}

\author{Chenglong Qian, Zhaohong Xiang, Zhuoran Wu and Hongbin Sun*
\thanks{*This work was supported by National Natural Science Foundation of China (No. 61790563).}
\thanks{The authors are with Xi'an Jiaotong University, Xi'an, Shaanxi, P.R. China, 710049.
        {\tt\small (e-mail: qiancl5683@stu.xjtu.edu.cn and hsun@mail.xjtu.edu.cn)}} %
\thanks{Hongbin Sun is the corresponding author.}
    
}

\begin{document}

\maketitle
\thispagestyle{empty}
\pagestyle{empty}

\begin{abstract}

Simultaneous Localization and Mapping (SLAM) is considered to be an essential capability for intelligent vehicles and mobile robots. However, most of the current lidar SLAM approaches are based on the assumption of a static environment. Hence the localization in a dynamic environment with multiple moving objects is actually unreliable. The paper proposes a dynamic SLAM framework RF-LIO, building on LIO-SAM, which adds adaptive multi-resolution range images and uses tightly-coupled lidar inertial odometry to first remove moving objects, and then match lidar scan to the submap. Thus, it can obtain accurate poses even in high dynamic environments. The proposed RF-LIO is evaluated on both self-collected datasets and open Urbanloco datasets. The experimental results in high dynamic environments demonstrate that, compared with LOAM and LIO-SAM, the absolute trajectory accuracy of the proposed RF-LIO can be improved by 90\% and 70\%, respectively. RF-LIO is one of the state-of-the-art SLAM systems in high dynamic environments.

\end{abstract}

\section{INTRODUCTION}

\begin{figure}[tbph]
	\centering
	\includegraphics[width=0.9\linewidth,height=0.6\linewidth]{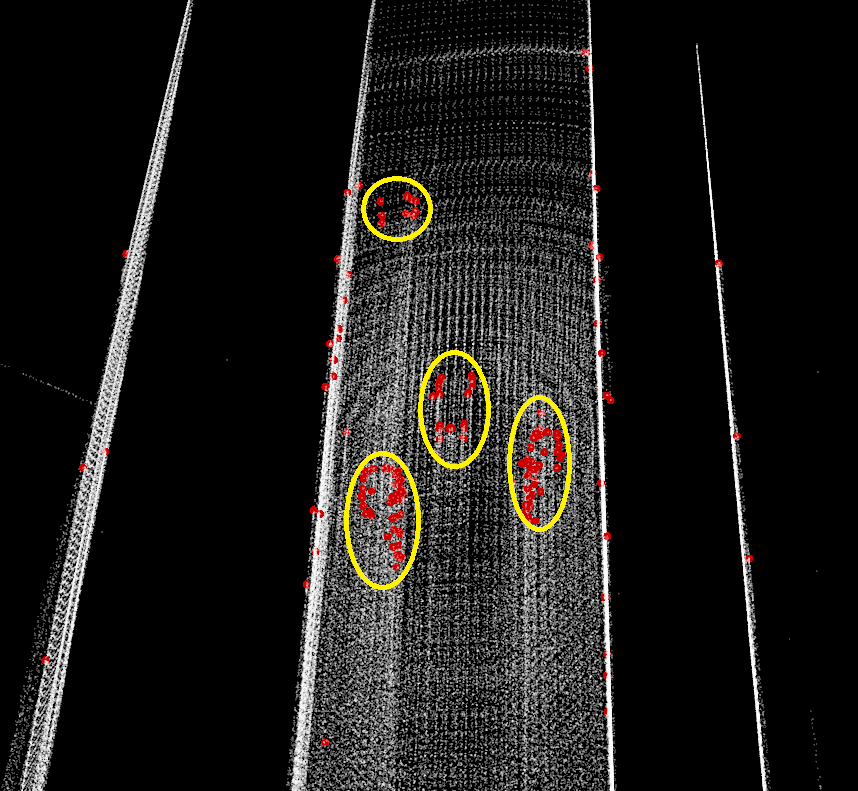}
	\caption{Distribution of feature points under the moving objects occlusion. The white points are the static environment. The red points are the edge feature points. The red points in the yellow circle fall on the moving vehicles.}
	\label{fig:moving-point}
\end{figure}

Robust and accurate localization is the premise for an intelligent vehicle or mobile robot. SLAM technology based on lidar can provide robust centimeter-level state estimation and a high-precision point cloud map without GPS, hence receives much attention recently. For example, LOAM \cite{Zhang2014} proposes a feature extraction strategy based on edge and plane, which has become the most widely used method for low-drift and real-time state estimation and mapping. On the basis of LOAM, LIO-SAM \cite{Shan2020} adopts tightly-coupled lidar inertial odometry (LIO) and keyframe strategy to further improve the processing speed and trajectory accuracy. Nevertheless, these mainstream SLAM systems and point cloud registration methods \cite{Segal2009,Biber2003} are based on the assumption of a static environment, i.e. there are no moving objects in the background. In the matter of fact, autonomous systems often work in a realistic environment with a lot of moving objects, such as vehicles, pedestrians, etc. Due to the sensor's field of view (FOV) is blocked, most of the feature points fall on the moving objects instead of the static map, as illustrated in Fig. \ref{fig:moving-point}. The previous approaches based on the static environment assumption will fail. Moreover, it is also challenging to extract road markers, traffic signs, and other critical static features in the point cloud map, as ghost tracks of moving objects may occlude them \cite{Pagad2020}. Therefore, it is essential to improve the performance of SLAM in dynamic environments.

A naive solution for dynamic SLAM is to build a map containing only static objects, i.e. to remove the moving object points from the point cloud map. For example, Suma++ \cite{Chen2019} first obtain the accurate poses of multiple measurements by scan-matching and then iteratively update the state (i.e. static or dynamic) of a map voxel by fusing the semantic information. However, this premise is often not tenable in a high dynamic environment because a large number of moving objects will make the existing scan-matching method ineffective. Hence, these methods fall into a chicken-and-egg problem: moving object points removal relies on accurate pose, while accurate pose can not be obtained when the point cloud map contains moving objects. The above problem makes these SLAM methods can not achieve good results in high dynamic environments. In addition, deep-learning methods relies heavily on training data, which also limits the application of Suma++ from one scene to other scenes.

This paper proposes a new removal-first tightly-coupled lidar inertial odometry framework, i.e. RF-LIO, to solve the SLAM problem in high dynamic environments. The removal-first refers to that the proposed RF-LIO first removes moving object without an accurate pose and then employ scan-matching. When a new scan arrives, RF-LIO does not immediately perform scan-matching to obtain an accurate pose, as it is easily affected by the dynamic environment. Instead, we use the tightly-coupled inertial measurement unit (IMU) odometry to obtain a rough initial state estimation. Then, RF-LIO can preliminarily remove the moving points in the environment by using the adaptive resolution range image. After preliminary moving points removal, RF-LIO uses scan-matching to obtain a relatively more accurate pose. Through these iterative removal and scan-matching steps, RF-LIO can finally obtain accurate poses in high dynamic environments. We use the removal rate of moving objects and absolute trajectory accuracy to evaluate RF-LIO on both self-collected datasets and open Urbanloco datasets \cite{Wen2020}. The experiments demonstrate that the average moving objects removal rate of RF-LIO is 96.1\% and the absolute trajectory accuracy of RF-LIO can be improved by 90\% and 70\%, compared with LOAM and LIO-SAM, respectively. 

\section{RELATED WORKS}

Moving objects such as pedestrians and vehicles widely exist in real-world scenes. Therefore, most state-of-the-art SLAM approaches that initially designed in static environments cannot handle these severe dynamic scenes \cite{Yu2018}. To address this problem, moving objects need to be recognized from the background and removed. The related approaches can be summarized as follows:

\textbf{Statistical approaches:} These approaches first calculate a geometric model, and then use a statistical method, such as Random Sample Consensus (RANSAC) \cite{Fischler1981}, to remove the feature points that do not conform to the geometric model. However, these approaches will fail if the moving features are in the majority. 

\textbf{Voxel-based approaches:} These methods are based on voxel ray casting \cite{Chen2016,Gehrung2017,Schauer2018}. Voxel-based approaches construct a huge voxel map and use the simple prior knowledge: when a lidar beam hits a voxel, the voxels along the way must be empty. But the premise of using these methods is to have very accurate localization information, which is contrary to the original intention of dynamic SLAM. In addition, these methods consume a lot of memory and computing resources. Even if the latest method \cite{Pagad2020} uses deep-learning and GPU acceleration, it can only maintain a maximum octree depth of 16 and voxel size of 0.3 meters.

\textbf{Visibility-based approaches:} These visibility-based approaches \cite{Ambrus2014,Xiao2015} do not need to maintain a vast voxel map. Compared with voxel-based approaches, these methods can build a larger map with higher computational efficiency. These methods are based on the principle that when there is a closer point in a narrow FOV, the farther point will be occluded. On the contrary, if we observe the map's far point, then the corresponding closer point in the query scan must be moving.

\textbf{Segmentation-based approaches:} DynaSLAM \cite{Bescos2018} use semantic information to assist moving objects detection. These methods often need to be combined with other traditional methods. Because even the object is identified as a vehicle, we can not judge whether it is stopping or driving. For example, DS-SLAM \cite{Yu2018} uses a statistical method to calculate the motion consistency of feature points to select moving points and fixed points. The proportion of moving points in the same object is used to judge whether the object is moving or stationary. Moreover, segmentation-based approaches rely heavily on supervised labels and training data and are limited by the accuracy and category of segmentation.

\section{ REMOVAL-FIRST LIDAR INERTIAL ODOMETRY}
\subsection{System Overview}

\begin{figure*}[tbph]
	\centering
	\includegraphics[width=0.8\linewidth,height=0.4\linewidth]{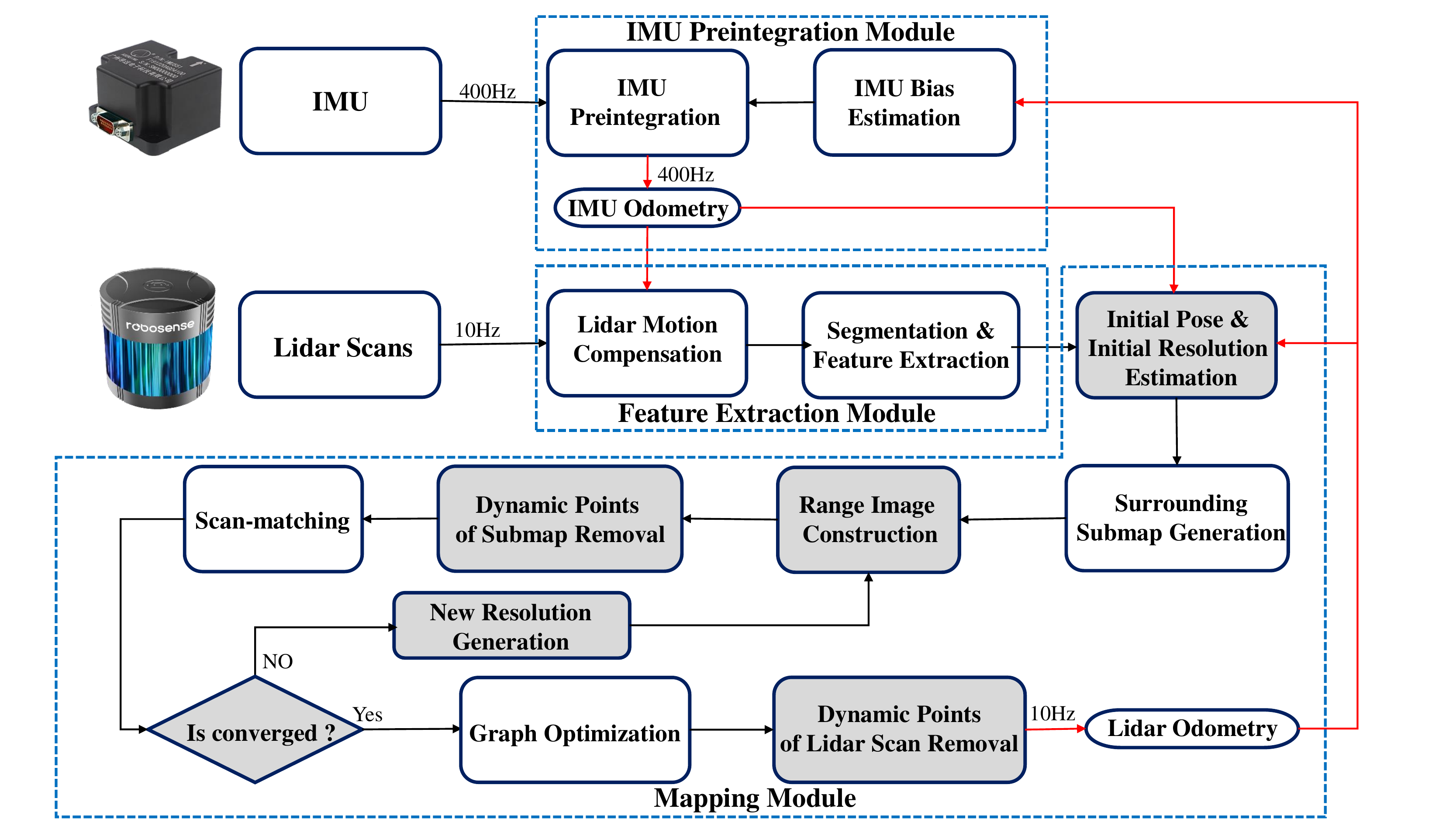}
	\caption{Overall framework of RF-LIO. The IMU preintegration module is used to infer the system motion and generate IMU odometry. The feature extraction module compensates the motion distortion of point cloud and extracts feature points by evaluating the roughness of points. The mapping module outputs a refined pose estimation and a global 3D mapping, and removes the moving objects from the point cloud map.
	}
	\label{fig:systeam}
\end{figure*}

Fig. \ref{fig:systeam} shows an overall framework of RF-LIO, which consists of three major modules: IMU preintegration, feature extraction, and mapping. First of all, the IMU preintegration module is used to infer the system motion and generates IMU odometry. Then, the feature extraction module compensates the motion distortion of the point cloud \cite{LeGentil2019}. The edge and plane features are extracted by
evaluating the roughness of points.

The mapping module is the critical module of our proposed approach. To achieve the moving objects removal-first without an accurate pose, there are several key steps: (\romannumeral1) The initial pose is obtained by IMU odometry. Then the error between IMU preintegration and scan-matching is used to determine an initial resolution (i.e. how many angles of FOV does each pixel correspond to). (\romannumeral2) RF-LIO use this initial resolution to construct the range image from the current lidar scan and the corresponding submap separately. (\romannumeral3) Through comparing their visibility, the main moving points of the submap are removed. (\romannumeral4) RF-LIO match the lidar scan to the submap and judge whether scan-matching is convergent. If it is convergent, after graph optimization, the final fine resolution is used to remove the remaining moving points in the current keyframe. otherwise, a new resolution will be generated, and steps (\romannumeral2), (\romannumeral3), and (\romannumeral4) will be repeated. 

\subsection{IMU Preintegration and Initial Pose} \label{section:A}
We first denote the world frame as $\boldsymbol{W}$ and the IMU frame, which coincides with the robot body frame as $\boldsymbol{B}$. The state of the system can be represented as follows:
\begin{equation}
\boldsymbol{x}=\left[\boldsymbol{R}^{\boldsymbol{T}}, \boldsymbol{p}^{\boldsymbol{T}}, \boldsymbol{v}^{\boldsymbol{T}}, \boldsymbol{b}^{\boldsymbol{T}}\right]^{\boldsymbol{T}} \label{eq:state}
\end{equation}
where $\boldsymbol{R} \in \mathrm{SO}(3)$ is the rotation matrix, $\boldsymbol{p} \in \mathbb{R}^{3}$ is the position vector, $\boldsymbol{v}$ is the speed vector, and $\boldsymbol{b}$ is the IMU bias vector.  $\boldsymbol{b}$ consists of slowly varying accelerometer bias $\boldsymbol{b}^\mathrm{a}$ and gyroscope bias $\boldsymbol{b}^\mathrm{g}$. The IMU measurement model can be expressed as follows: 
\begin{align}
	\hat{\boldsymbol{\omega}}_\mathrm{B}(t)&=\omega_\mathrm{B}(t)+\boldsymbol{b}^{\mathrm{g}}(t)+\boldsymbol{\eta}^{\mathrm{g}}(t) \\		\hat{\boldsymbol{a}}_\mathrm{B}(t)&=\boldsymbol{R}_{\mathrm{WB}}^\mathrm{T}(t)\left(\boldsymbol{a}_\mathrm{B}(t)-\boldsymbol{g}_\mathrm{W}\right)+\boldsymbol{b}^{\mathrm{a}}(t)+\boldsymbol{\eta}^{\mathrm{a}}(t)
\end{align}
Where $\hat{\boldsymbol{\omega}}_\mathrm{B}$ and $\hat{\boldsymbol{a}}_\mathrm{B}$ are the raw IMU measurements in $\boldsymbol{B}$ and are affected by the bias $\boldsymbol{b}$ and white noise $\boldsymbol{\eta}$.

When IMU measurements come, we apply the IMU preintegration method proposed in \cite{Forster2015} to obtain the initial pose of the current keyframe $k+1$ from the previous keyframe $k$: 
\begin{align}
	\boldsymbol{R}_{\mathrm{WB}}^{k+1} &=\boldsymbol{R}_{\mathrm{WB}}^{k} \Delta \boldsymbol{R}_{k, k+1} \operatorname{Exp}\left(\left(\boldsymbol{J}_{\Delta R}^\mathrm{g} \boldsymbol{b}^\mathrm{g}_{k}\right)\right) \\
	\boldsymbol{v}_{\mathrm{B}}^{k+1} &=\boldsymbol{v}_{\mathrm{B}}^{k}+\boldsymbol{g}_{\mathrm{W}} \Delta t_{k, k+1} \notag\\
	&+\boldsymbol{R}_{\mathrm{WB}}^{k}\left(\Delta \boldsymbol{v}_\mathrm{B}^{k, k+1}+\boldsymbol{J}_{\Delta v}^\mathrm{g} \boldsymbol{b}^\mathrm{g}_{k}+\boldsymbol{J}_{\Delta v}^\mathrm{a} \boldsymbol{b}^\mathrm{a}_{k}\right) \\
	\boldsymbol{p}_{\mathrm{B}}^{k+1} &=\boldsymbol{p}_{\mathrm{B}}^{k}+\boldsymbol{v}_{\mathrm{B}}^{k} \Delta t_{k, k+1}+\frac{1}{2} \boldsymbol{g}_{\mathrm{W}} \Delta t_{k, k+1}^{2} \notag\\
	&+\boldsymbol{R}_{\mathrm{WB}}^{k}\left(\Delta \boldsymbol{p}_{k, k+1}+\boldsymbol{J}_{\Delta p}^{g} \boldsymbol{b}^\mathrm{g}_{k}+\boldsymbol{J}_{\Delta p}^{a} \boldsymbol{b}^\mathrm{a}_{k}\right)
\end{align}
Where the Jacobians $\boldsymbol{J}_{(\cdot)}^{a}\boldsymbol{b}^\mathrm{a}$ and $\boldsymbol{J}_{(\cdot)}^{g}\boldsymbol{b}^\mathrm{g}$ represent a first-order approximation of the effect of changing the biases without explicitly recomputing the preintegration.

\subsection{IMU Preintegration Error and Initial Resolution} \label{section:B}

In the process of using IMU measurements to infer system motion, there will inevitably be a deviation from the ground truth, which makes the query scan points to the corresponding map points are ambiguous. To solve this problem, Palazzolo and Stachniss \cite{Palazzolo2018} proposed a window-based method (i.e. not pixel-to-pixel, but pixel-to-window comparison). A more convenient method was proposed in Removert \cite{Kim2020} that uses multiple range images having different resolutions. However, Removert uses fixed resolutions, as it is based on accurate localization information. But RF-LIO needs to remove the dynamic points before scan-matching (i.e. without an accurate pose). Hence we use the pose error between IMU preintegration and scan-matching to generate the initial resolution dynamically.

When scan-matching is performed, the translation and orientation errors of IMU preintegration can be computed as follows:
\begin{align}
	\boldsymbol{E}_\mathrm{R}^{k} &=\text{Log} \left(\left(\Delta \boldsymbol{R}_{k-1,k} \operatorname{Exp}\left(\boldsymbol{J}_{\Delta R}^\mathrm{g} \boldsymbol{b}^\mathrm{g}_{k-1}\right)\right)^{T} \boldsymbol{R}_{\mathrm{BW}}^{k-1} \boldsymbol{R}_{\mathrm{WB}}^{k}\right)  \label{eq:e_R}\\
	\boldsymbol{E}_\mathrm{v}^{k} &=\boldsymbol{R}_{\mathrm{BW}}^{k-1}\left(\boldsymbol{v}_{\mathrm{B}}^{k}- \boldsymbol{v}_{\mathrm{B}}^{k-1}-\boldsymbol{g}_{\mathrm{W}} \Delta t_{k-1,k}\right) \notag \\
	&-\left(\Delta \boldsymbol{v}_{k-1,k}+\boldsymbol{J}_{\Delta v}^\mathrm{g} \boldsymbol{b}^\mathrm{g}_{k-1}+\boldsymbol{J}_{\Delta v}^\mathrm{a} \boldsymbol{b}^\mathrm{a}_{k-1}\right) \label{eq:e_v}\\
	\boldsymbol{E}_\mathrm{p}^{k} &=\boldsymbol{R}_{\mathrm{BW}}^{k-1}\left(\boldsymbol{p}_{\mathrm{B}}^{k}-\boldsymbol{p}_{\mathrm{B}}^{k-1}-\boldsymbol{v}_{\mathrm{B}}^{k-1} \Delta t_{k-1,k}-\frac{1}{2} \boldsymbol{g}_{\mathrm{W}} \Delta t_{k-1,k}^{2}\right) \notag \\
	&-\left(\Delta \boldsymbol{p}_{k-1,k}+\boldsymbol{J}_{\Delta p}^\mathrm{g} \boldsymbol{b}^\mathrm{g}_{k-1}+\boldsymbol{J}_{\Delta p}^{a} \boldsymbol{b}^\mathrm{a}_{k-1}\right) \label{eq:e_p}\\
	\boldsymbol{E}_\mathrm{b}^{k} &=\boldsymbol{b}_{k}-\boldsymbol{b}_{k-1} \label{eq:e_b}
\end{align}

Through (\ref{eq:e_R}), (\ref{eq:e_v}), (\ref{eq:e_p}), and (\ref{eq:e_b}), we can get the pose error of the previous keyframe $k$. However, before scan-matching, the IMU preintegration error of the current keyframe $k+1$ can not be obtained by using the above method. To get the error of the current keyframe $k+1$, we use the error transfer relation of the nonlinear system:
\begin{equation}
	\delta \boldsymbol{X}_{k}=\delta \boldsymbol{X}_{k-1}+\delta \dot{\boldsymbol{X}}_{k-1} \Delta t
\end{equation}

We are only interested in the $\delta \boldsymbol{\theta}$ and $\delta \boldsymbol{p}$ of $\delta \boldsymbol{X}$, hence our approach can be written as follows:
\begin{equation}
	\left[\begin{array}{c}
		\vspace{1ex} 
		\delta \boldsymbol{\theta}^{k+1} \\	
		\vspace{1ex} 
		\delta \boldsymbol{p}^{k+1} \\	
	\end{array}\right]=\boldsymbol{A}\left[\begin{array}{c}	
		\vspace{1ex} 
		\delta \boldsymbol{\theta}^{k} \\
		\vspace{1ex} 
		\delta \boldsymbol{p}^{k} \\	
		\vspace{1ex} 
		\delta \boldsymbol{v}^{k} \\		
		\vspace{1ex} 
		\delta \boldsymbol{b}_{k}^\mathrm{a} \\		
		\vspace{1ex} 
		\delta \boldsymbol{b}_{k}^\mathrm{g}
	\end{array}\right]+\boldsymbol{B}\left[\begin{array}{c}	
		\vspace{1ex} 
		\boldsymbol{\eta}^\mathrm{g} \\	
		\vspace{1ex} 
		\boldsymbol{\eta}^\mathrm{a} \\	
		\vspace{1ex} 	
		\boldsymbol{\eta}_\mathrm{b^\mathrm{g}} \\	
		\vspace{1ex} 	
		\boldsymbol{\eta}_\mathrm{b^\mathrm{a}}
	\end{array}\right]
\end{equation}
\begin{equation}
	\boldsymbol{A}=\left[\begin{array}{ccccc}
		\vspace{1ex} 
		\boldsymbol{I}-[\boldsymbol{\omega}]_{\times} \Delta t &\boldsymbol{0} &  \boldsymbol{0} & \boldsymbol{0} & -\boldsymbol{I} \Delta t \\
		0 &\boldsymbol{I} &  \boldsymbol{I} \Delta t & 0 & 0
	\end{array}\right]
\end{equation}


\begin{equation}
	\boldsymbol{B}=\left[\begin{array}{cccc}
		\vspace{1ex}
		\boldsymbol{I} \Delta t& \boldsymbol{0}  & \boldsymbol{0} & \boldsymbol{0} \\
		\boldsymbol{0} & \boldsymbol{0}  & \boldsymbol{0} & \boldsymbol{0}
	\end{array}\right]
\end{equation}

By predicting the IMU odometry localization error, we can obtain the initial resolution of the range image. We use the following empirical formula to convert translation and orientation error into resolution:
\begin{equation}
	\boldsymbol{r}=\alpha\boldsymbol{\delta p} + \boldsymbol{\delta \theta} \label{eq:resolution1}
\end{equation}
Where $\alpha$ is a value between 0 and 1, which is used to balance the contribution of translation error and orientation error to resolution.

To balance the real-time performance and removal rate, and to avoid the influence caused by the wrong prediction resolution, we set a minimum resolution $\boldsymbol{r}_0$ and appropriately enlarge the predicted resolution $\boldsymbol{r}$. The final initial resolution $\boldsymbol{r}_{\mathrm{f}}$ is defined as follows:
\begin{equation}
	\boldsymbol{r}_{\mathrm{f}}=max(\beta \boldsymbol{r}, \boldsymbol{r}_0) \label{eq:resolution2}
\end{equation}
Where $\beta$ is a coefficient greater than 1. $\boldsymbol{r}_0 = \frac{\text{vertical FOV}}{\text{vertical rays}}$, which can effectively avoid empty pixel values in range image.

\subsection{Range Image Construction and Moving Points Removal} \label{section:C}

We first define $\boldsymbol{F}_{k+1}$ as the current keyframe scan and $\boldsymbol{M}_k$ as a corresponding submap, which is a point cloud map created by sliding window method around $\boldsymbol{F}_{k+1}$. In addition, to balance the removal rate of moving points and the real-time performance, we use a full query scan to compare with the feature submap. This is because a feature submap with multiple keyframes has a similar density to the full query scan and has fewer points than a full submap. Then, we divide the point cloud into two mutually exclusive subsets: dynamic points set $(\cdot)^\mathrm{D}$, and static points set $(\cdot)^\mathrm{S}$. Formally, the aforementioned problem is expressed as: 
\begin{equation}
	\boldsymbol{M}=\boldsymbol{M}^\mathrm{D} \cup \boldsymbol{M}^\mathrm{S}  \qquad
	\boldsymbol{F}=\boldsymbol{F}^\mathrm{D} \cup \boldsymbol{F}^\mathrm{S}
\end{equation}
while $(\cdot)^\mathrm{D} \cap (\cdot)^\mathrm{S}=\varnothing$.

\begin{figure}[tbph]
	\centering
	\includegraphics[width=1.0\linewidth,height=0.5\linewidth]{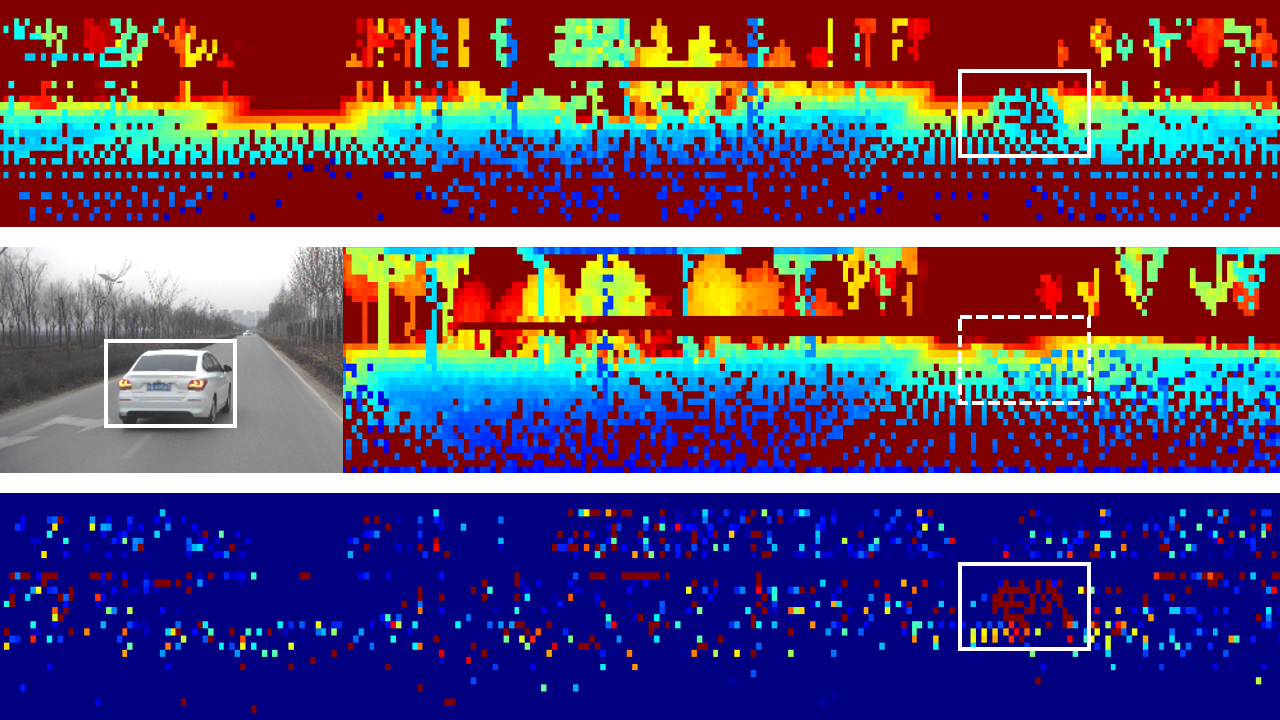}
	\caption{Adaptive resolution range image. The top row is the range image of the current scan $I_{k+1}^\mathrm{F}$. The middle row is the range image of the surrounding submap $I_{k}^\mathrm{M}$. The bottom row is the difference range image $I_{k+1}^{\text {Diff }}$. The color of a pixel indicates the distance of a pixel's corresponding 3D point to the sensor keyframe $k+1$; blue indicates the closer distance, and red indicates the farther distance. Therefore, the red pixels in the bottom range image represent the moving points in the scan.}
	\label{fig:range image}
\end{figure}

Sharing the analogous philosophy to existing approaches \cite{Kim2020,Palazzolo2018}, we use the projected range image's visibility to perform moving points recognition. As shown in Fig. \ref{fig:range image}, The pixel $(i,j)$ value $\boldsymbol{I}_{k+1, i j}$ of the range image is defined as the distance of a point $\boldsymbol{p} \in \mathbb{R}^{3}$ to the $k+1$th keyframe's local coordinate system $B_{k+1}$:
\begin{equation}
	\boldsymbol{I}_{k, i j}^\mathrm{M}=\min _{\boldsymbol{p} \in \boldsymbol{P}_{i j}^\mathrm{M}} \text{dist}(\boldsymbol{p}) \qquad
	\boldsymbol{I}_{k+1, i j}^\mathrm{F}=\min _{\boldsymbol{p} \in \boldsymbol{P}_{i j}^\mathrm{F}} \text{dist}(\boldsymbol{p})
\end{equation}
Where the range image size is determined by the given resolution and the lidar's horizontal and vertical FOV range. $\boldsymbol{P}_{i j}^\mathrm{ \text{M or F}}$ is the spherical coordinate of the point $\boldsymbol{P}^\mathrm{ \text{M or F}}$ (i.e. azimuth angle $i$ and elevation angle $j$ ).

Then, the visibility of submap points and scan points are calculated via their matrix element-wise subtraction as:
\begin{equation}
	I_{k+1}^{\text {Diff }}=I_{k+1}^\mathrm{F}-I_{k}^\mathrm{M}
\end{equation}

We determine a point $\boldsymbol{p} \in \boldsymbol{P}_{i j}^\mathrm{ \text{M or F}}$ is dynamic if its corresponding pixel value of $I_{k+1,ij}^{\text {Diff }}$ is larger than an adaptive threshold $\tau$. Finally, the dynamic points are defined as: 
\begin{align}
	\boldsymbol{M}_{k}^\mathrm{D}&=\left\{\boldsymbol{M}_{k} \mid \boldsymbol{I}_{k+1}^{\text {Diff }}>\tau\right\} \\
	\boldsymbol{F}_{k+1}^\mathrm{D}&=\left\{\boldsymbol{F}_{k+1} \mid \boldsymbol{I}_{k+1}^{\text {Diff }}<-\tau\right\} \\
	\tau&= \gamma \text{dist}(\boldsymbol{p})
\end{align}
Where $\gamma$ is the sensitivity with respect to the point's distance. We suggest setting $\gamma$ to the maximum of the final resolution shown in Fig. \ref{sub.3}, which can effectively avoid mistaking the static points for the dynamic points due to the wrong prediction resolution.

\subsection{Lidar Odometry and New Resolution} \label{section:D}
The lidar odometry is used to estimate the sensor motion between two consecutive scans. Various scan-matching methods, such as \cite{Segal2009} and \cite{Biber2003}, can be utilized for this purpose. We opt to use the method proposed in LOAM \cite{Zhang2014}, as a lot of related works \cite{Shan2018,Qin2019} has proved that it has an excellent and robust performance in various challenging environments. Then the edge and plane features are extracted for each new lidar scan by evaluating the roughness of points over a local area. We denote the edge and plane features from the keyframe $k+1$ as $\boldsymbol{F}^\mathrm{e}_{k+1}$ and $\boldsymbol{F}^\mathrm{p}_{k+1}$, respectively. Then we transform them from $\boldsymbol{B}_{k+1}$ to $\boldsymbol{W}$ and obtain
$\left\{{ }^{w} \boldsymbol{F}_{k+1}^\mathrm{e},{ }^{w} \boldsymbol{F}_{k+1}^\mathrm{p}\right\}$. This initial transformation is obtained by IMU preintegration (see Sec. \ref{section:A} for details). For each feature point in $\left\{{ }^{w} \boldsymbol{F}_{k+1}^\mathrm{e},{ }^{w} \boldsymbol{F}_{k+1}^\mathrm{p}\right\}$, we can find its correspondence feature points in $\left\{\boldsymbol{M}_{k}^\mathrm{e}, \boldsymbol{M}_{k}^\mathrm{p}\right\}$ via nearest neighbor search. Then the distance between a feature point and its corresponding edge or planar patch can be calculated using the following equations:
\begin{equation}
	\boldsymbol{d}^\mathrm{e}_{k}=\frac{\left|\left(\boldsymbol{p}_{k+1, i}^\mathrm{e}-\boldsymbol{p}_{k, j}^\mathrm{e}\right) \times\left(\boldsymbol{p}_{k+1, i}^\mathrm{e}-\boldsymbol{p}_{k, l}^\mathrm{e}\right)\right|}{\left|\boldsymbol{p}_{k, j}^\mathrm{e}-\boldsymbol{p}_{k, l}^\mathrm{e}\right|} 
\end{equation}
\begin{equation}
	\boldsymbol{d}^\mathrm{p}_{k}=\frac{\left|\begin{array}{c} \left(\boldsymbol{p}_{k+1, i}^\mathrm{p}-\boldsymbol{p}_{k, j}^\mathrm{p}\right)\\
			\left(\boldsymbol{p}_{k, j}^\mathrm{p}-\boldsymbol{p}_{k, l}^\mathrm{p}\right) \times\left(\boldsymbol{p}_{k, j}^\mathrm{p}-\boldsymbol{p}_{k, m}^\mathrm{p}\right)\end{array} \right|}{\left|\left(\boldsymbol{p}_{k, j}^\mathrm{p}-\boldsymbol{p}_{k, l}^\mathrm{p}\right) \times\left(\boldsymbol{p}_{k, j}^\mathrm{p}-\boldsymbol{p}_{k, m}^\mathrm{p}\right)\right|}
\end{equation}
Where $i, j, l$ and $m$ are the feature indices. $\boldsymbol{p}_{k+1, i}^\mathrm{e} \in { }^{w} \boldsymbol{F}_{k+1}^\mathrm{e}$ is a edge feature point, and $\boldsymbol{p}_{k, j}^\mathrm{e}, \boldsymbol{p}_{k, l}^\mathrm{e} \in  \boldsymbol{M}_{k}^\mathrm{e} $ are corresponding edge-line points. $\boldsymbol{p}_{k+1, i}^\mathrm{p} \in { }^{w} \boldsymbol{F}_{k+1}^\mathrm{p}$ is a plane feature point, and $\boldsymbol{p}_{k, j}^\mathrm{p}, \boldsymbol{p}_{k, l}^\mathrm{p}, \boldsymbol{p}_{k, m}^\mathrm{p} \in  \boldsymbol{M}_{k}^\mathrm{p} $ are corresponding planar patch points. Finally. we can obtain the transformation by solving the optimization problem:
\begin{equation}
\begin{aligned}
	\min\left\{ \boldsymbol{\Delta T_{k,k+1}} \right\} =  
	\min _{\Delta \boldsymbol{T}_{k,k+1}}\left\{\sum_{\boldsymbol{p}_{k+1, i}^\mathrm{e} \in {}^{w} \boldsymbol{F}_{k+1}^\mathrm{e}} \boldsymbol{d}^\mathrm{e}_{k}+\sum_{\boldsymbol{p}_{k+1, i}^\mathrm{p} \in { }^{w} \boldsymbol{F}_{k+1}^\mathrm{p}} \boldsymbol{d}^\mathrm{p}_{k}\right\}
\end{aligned}
\end{equation}

When the scan-matching is convergent, we can get the relative transformation $\Delta \boldsymbol{T}_{k,k+1}$ between the two keyframes $k$ and ${k+1}$. So the lidar odometry can be written as:
\begin{equation}
	\boldsymbol{T}_{k+1} = \boldsymbol{T}_{k}\Delta \boldsymbol{T}_{k,k+1}
\end{equation}

However, in a high dynamic environment with multiple moving objects, the lidar odometry will drift. As shown in Fig. \ref{fig:systeam}, we have preliminarily removed the moving points in the steps of Sec. \ref{section:B} and Sec. \ref{section:C}, but it still can not guarantee the reliability of scan-matching. Hence we need to judge its convergence. We describe and implement a naive but effective approach, which is judged by the Euler distance-based of the edge points $\boldsymbol{F}^\mathrm{e}_{k+1}$. It can be naturally embedded into the our scan-matching method without extra computation. Besides, compared with using all scan points, the sparsity of edge points can ensure that the method is robust enough. We also note that our framework can be compatible with other methods, such as \cite{Chen2020} and \cite{Pomerleau2014}, but these methods need other time to calculate. Our experiments show that our approach is effective enough. It is summarized in Alg. \ref{Alg:score}.

\begin{algorithm}[tbph]
	\caption{ Convergence Score Calculation} 
	\label{Alg:score}
	\hspace*{0.02in} {\bf Input:} \\
	\hspace*{2em} Edge points of current scan: $ \{\boldsymbol{F}^\mathrm{e}_{k+1}\}$ \\
	\hspace*{2em} Edge points of corresponding submap: $\{ \boldsymbol{M}^\mathrm{e}_k\}$ \\
	\hspace*{0.02in} {\bf Output:}  \\
	\hspace*{2em} Convergence Score \\
	\hspace*{0.02in} {\bf Procedure:}
	\begin{algorithmic}[1] 
		\State Transform  $ \{\boldsymbol{F}^\mathrm{e}_{k+1}\}$ from body frame $ \boldsymbol{B}_{k+1}$ to world frame $ \boldsymbol{W}$ %
		\State Initialize $\text{score}=0$ and number $n_{\text{score}}=0$
		\For{$\boldsymbol{p}^\mathrm{e}_{i}$ in $\{\boldsymbol{F}^\mathrm{e}_{k+1}\}$} %
		\State Find its nearest neighbours in $\{ \boldsymbol{M}^\mathrm{e}_k\}$ by KDtree,
		\hspace*{1.2em} then calculate their distance
		\If{$\text{distance} \leqslant \tau_\mathrm{D}$} %
		\State $\text{score}+=\text{distance}$ 
		\State $n_{\text{score}}+=1$
		\EndIf
		\EndFor
		\If{$n_{\text{score}}>0$} %
		\State $\text{score}=\frac{\text{score}}{n_{\text{score}}}$ 
		\Else
		\State $\text{score}=\infty$
		\EndIf
	\end{algorithmic}
\end{algorithm}
Where $\tau_\mathrm{D}$ is a threshold used to remove outliers that are too far away. When $\text{score} < \text{score}_0$, it is judged to be convergent, otherwise it is not convergent.

When the scan-matching is not convergent, RF-LIO use (\ref{eq:resolution1}) and (\ref{eq:resolution2}) to generate a new resolution better than the initial resolution based on the pose error of lidar Odometry, then removes the dynamic points again and repeats the steps of Set. \ref{section:D}.

\section{EXPERIMENTS}
\subsection{Experimental Setup}

\begin{table}[htbp]
	\centering 
	\caption{Parameters Used in RF-LIO for All Experiments.}
	\label{table:Parameter}  
	\begin{tabular}{cccccc}
		\hline
		\textbf{Parameter} & $\alpha$ & $\beta$ & $r_0$ & $\gamma$ & $\text{score}_0$\\
		\hline
		\textbf{Value} & 0.1  & 2.0  & 0.02 & 0.02 & 0.25 \\
		\hline
	\end{tabular}
\end{table}

We evaluate our RF-LIO via a series of experiments and compare it with LOAM \cite{Zhang2014} and LIO-SAM \cite{Shan2020}. We note that both RF-LIO and LIO-SAM use the same feature extraction and loop closure detection methods, and both use GTSAM\cite{Kaess2012} to optimize the factor graph. The ablation experiment with LIO-SAM shows the effect of removal-first in RF-LIO. In all experiments, RF-LIO uses the same parameters shown in Table \ref{table:Parameter}. All the methods are implemented in C++ and executed on a computer with an Intel i7-10700k CPU using the robot operating system (ROS) \cite{Quigley2009} in Ubuntu Linux. For validation, we use self-collected datasets and open UrbanLoco datasets \cite{Wen2020}, where the Urbanloco datasets contain a large number of moving objects. The details of these datasets are shown in Table \ref{table: Data detail}. For a dataset with a few moving objects, we define it as a low dynamic dataset. For a dataset with a large number of moving objects, we define it as a high dynamic dataset. And medium dynamic datasets are between low dynamic datasets and high dynamic datasets. All the methods only use lidar and IMU without GPS. The GPS data is only used as the ground truth.

\begin{table}[htbp]
	\newcommand{\tabincell}[2]{\begin{tabular}{@{}#1@{}}#2\end{tabular}}
	\centering 
	\caption{Datasets Details}
	\label{table: Data detail}  
	\tabcolsep 0.03in
	\begin{tabular}{cccccc}
		\hline
		Dataset & Scans & \tabincell{c}{Trajectory \\
		Length (m)} & \tabincell{c}{Max\\ Speed (m/s)}& \tabincell{c}{Average \\ Speed (m/s)}& \tabincell{c}{ Dynamic\\ Level}\\
		\hline
		Urban & 16131& 6390.33 & 5.29 & 3.89 & Low \\
		Campus & 5883 & 1007.97 & 1.96 & 1.56 & Medium \\
		Suburban & 4198& 1890.44 & 6.20 & 4.64& Medium \\		
		CAMarketStreet & 14471 & 5690.98 & 13.44 & 7.07 & High \\
		CARussianHill & 15860 & 3570.38 & 10.27 & 4.98 & High \\
		\hline
	\end{tabular}
\end{table}

\subsection{Initial Resolution Analysis}

\begin{figure}[tbph]
	\centering
	\subfigure[Translation Error]{
		\label{sub.1}
		\includegraphics[width=1.0\linewidth,height=0.45\linewidth]{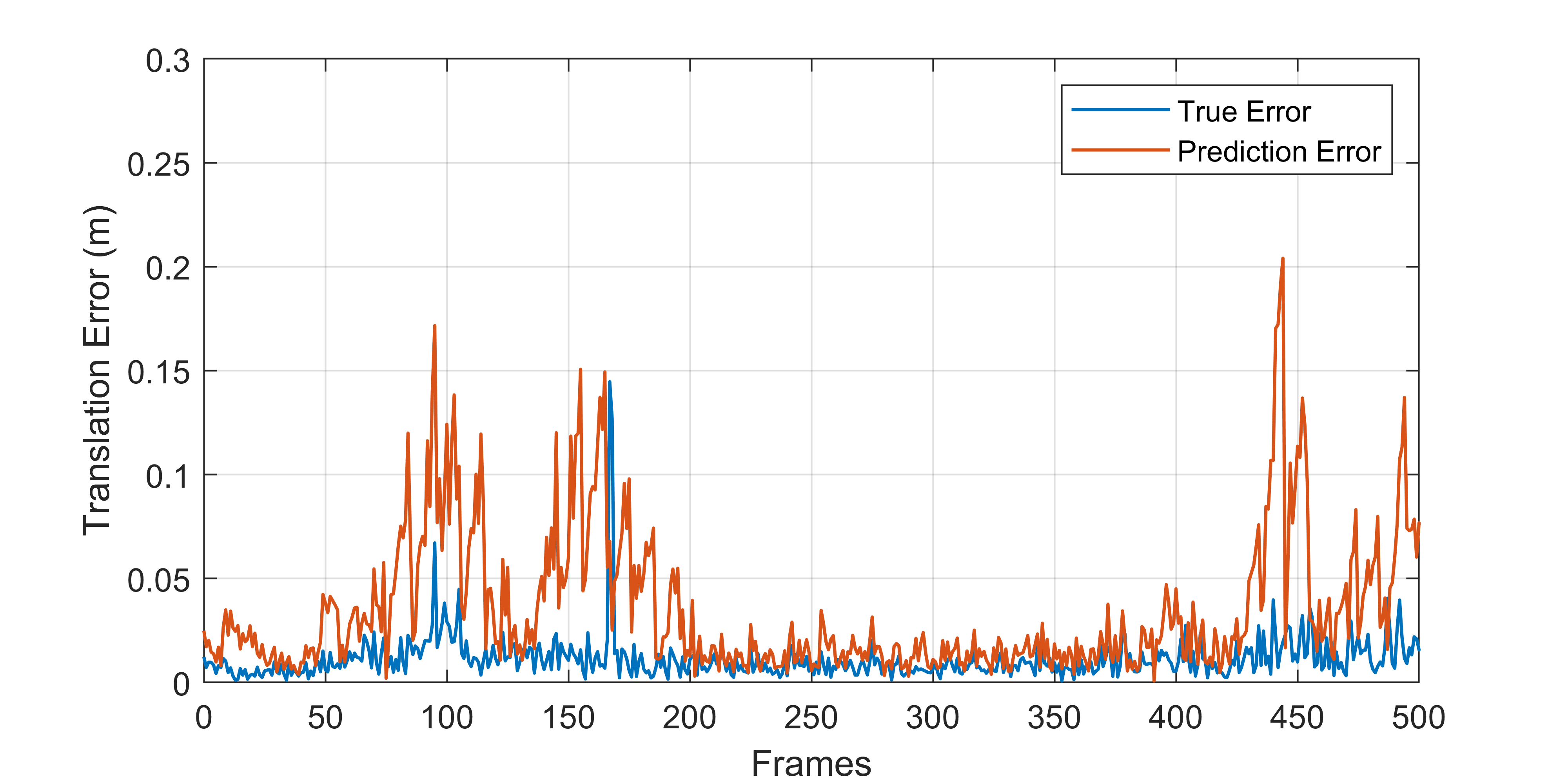}	
	}
	\subfigure[Orientation Error]{
		\label{sub.2}
		\includegraphics[width=1.0\linewidth,height=0.45\linewidth]{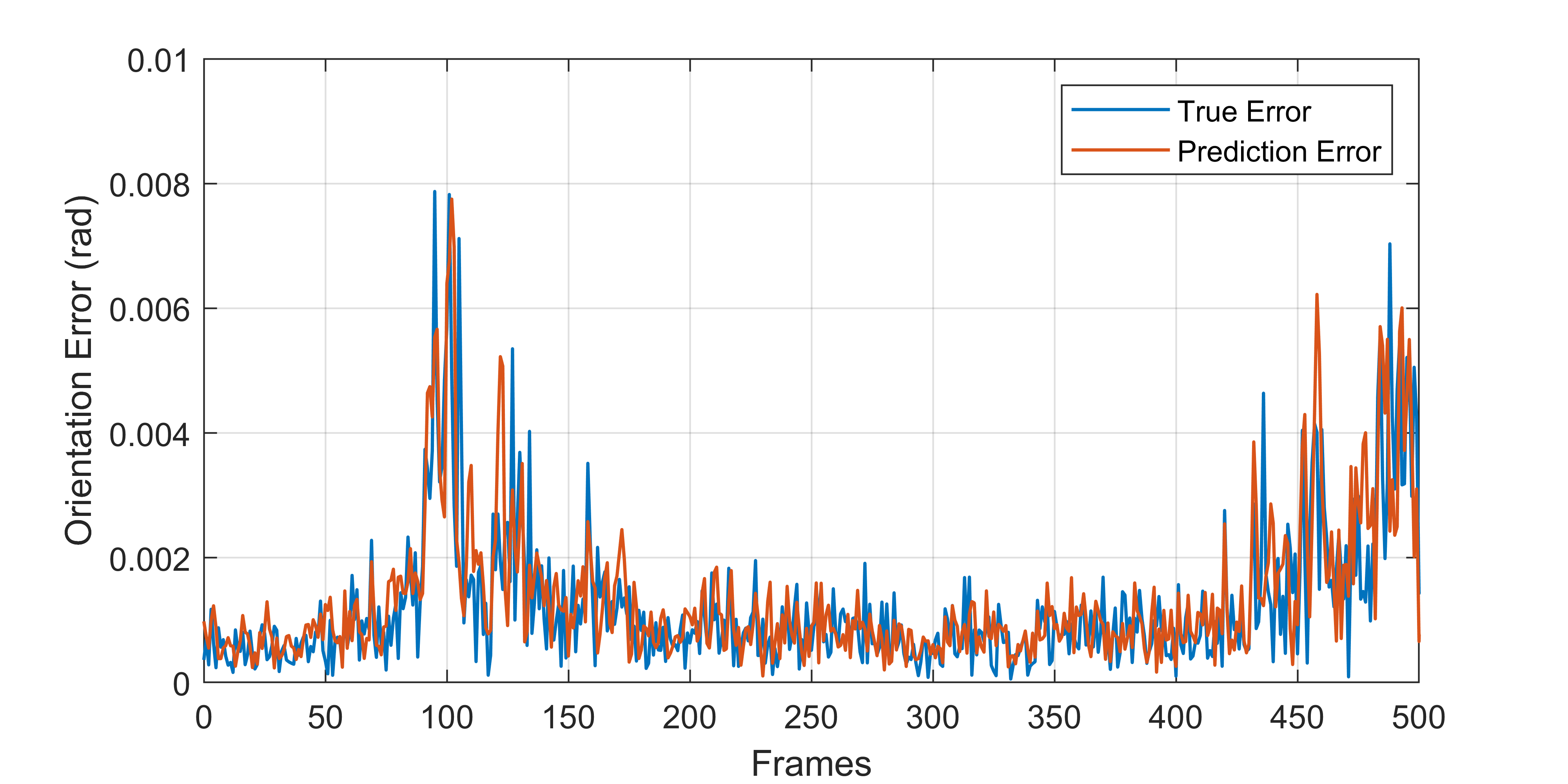}	
	}
	\subfigure[Initial Resolution]{
		\label{sub.3}
		\includegraphics[width=1.0\linewidth,height=0.45\linewidth]{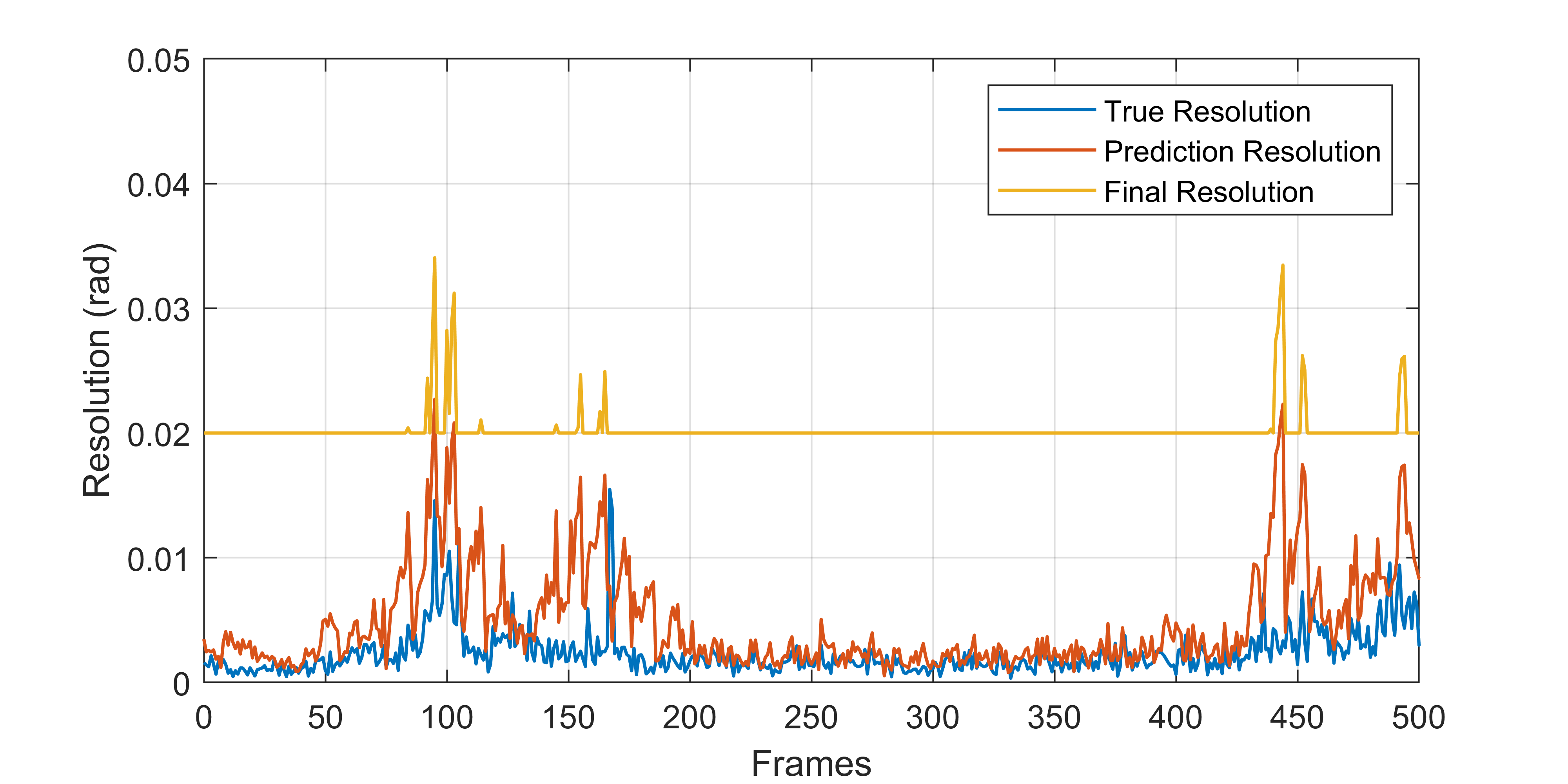}	
	}
	\caption{Translation error, orientation error and initial resolution analysis. (a) The translation error of IMU preintegration. (b) The orientation error of IMU preintegration. (c) The resolution curve. The true resolution and the prediction resolution are calculated by the true error and the prediction error, respectively. The final resolution is an appropriate amplification of the prediction resolution to reduce the impact of a wrong prediction.}
	\label{fig:initial resolution}
\end{figure}

Setting the correct initial resolution of the range image can effectively remove moving points and avoid the mistake of removing fixed points. Therefore, this experiment is designed to prove the correctness of the method proposed in Sec. \ref{section:B}. From Fig. \ref{sub.2}, we can see that the predicted orientation error is in good agreement with the true error. Because the translation is the double-integrating of acceleration, the predicted translation error is not as accurate as the predicted orientation error. Fig. \ref{sub.1} shows that the deviation between the predicted translation error and the true translation error is always within an acceptable range. As shown in Fig. \ref{sub.3}, the prediction resolution is almost consistent with the true resolution, and the final resolution is always bigger than the true resolution. It makes our method can effectively reduce the mistake of identifying moving points. 

\subsection{Moving Objects Removal Test}

\begin{figure}[tbph]
	\centering
	\subfigure[Test environment]{
		\label{fig:test:enviromnet}
		\includegraphics[width=0.8\linewidth,height=0.4\linewidth]{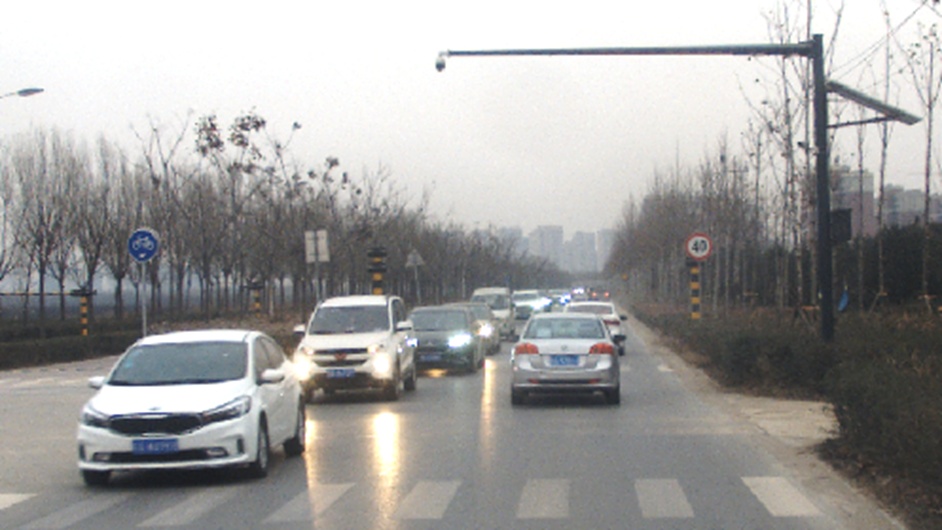}	
	}
	\subfigure[LIO-SAM]{
		\label{fig:LIO-SAM}
		\includegraphics[width=0.8\linewidth,height=0.4\linewidth]{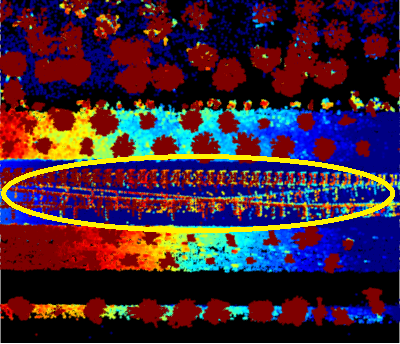}	
	}
	\subfigure[RF-LIO]{
		\label{fig:RF-LIO}
		\includegraphics[width=0.8\linewidth,height=0.4\linewidth]{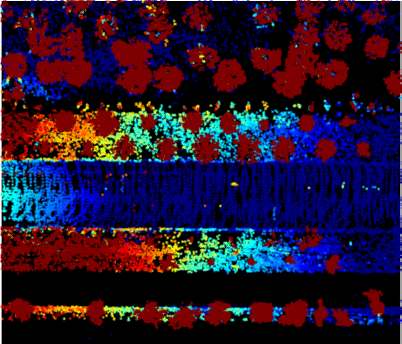}	
	}
	\caption{Mapping results of LIO-SAM and RF-LIO in the dynamic environment of the suburban dataset.}
	\label{fig:Removal Rate}
\end{figure}

In this test, the removal rate of moving points is used to evaluate RF-LIO. For this purpose, we collected three datasets with multiple moving objects in different environments. Fig. \ref{fig:test:enviromnet} shows a snapshot of the suburban dataset with many moving vehicles. RF-LIO has the same feature extraction method as LIO-SAM. Therefore, we use LIO-SAM as the evaluation standard to evaluate the moving points removal rate of RF-LIO. The maps obtained from LIO-SAM and RF-LIO are shown in Figs. \ref{fig:LIO-SAM} and \ref{fig:RF-LIO}, respectively. As is shown, LIO-SAM renders a lot of ghost tracks in the point cloud map. Compared with it, RF-LIO can get a purer map. We count the number of residual moving points in the maps of LIO-SAM and RF-LIO, respectively, and then calculate the removal rate. The results are shown in Table \ref{table: Removal Rate}. We can see that RF-LIO achieves an average removal rate of 96.1\% compared with LIO-SAM. We note that RF-LIO does not completely remove all the moving points. Because some moving points are too close to the ground ($distance < 0.5$, which is less than the threshold we set), and others are generated by lidar beams parallel to the ground, which makes it difficult to find the corresponding far points in the submap to remove them.

\begin{table}[htbp]
	\centering 
	\caption{Removal Rate of Moving Object Points}
	\label{table: Removal Rate}  
	\begin{tabular}{cccc}
		\hline
		Dataset & LIO-SAM & RF-LIO & Removal Rate\\
		\hline
		Urban &  85890   & 1803 & 97.9\% \\
		Campus & 134092 & 5766 & 95.7\% \\
		Suburban & 198113& 8898 & 95.5\% \\
		Average & 139365 & 5489 & 96.1\% \\
		\hline
	\end{tabular}
\end{table}

\subsection{Results on Low and Medium Dynamic Datasets}
\begin{figure}[tbph]
	\centering
	\includegraphics[width=1.0\linewidth,height=0.6\linewidth]{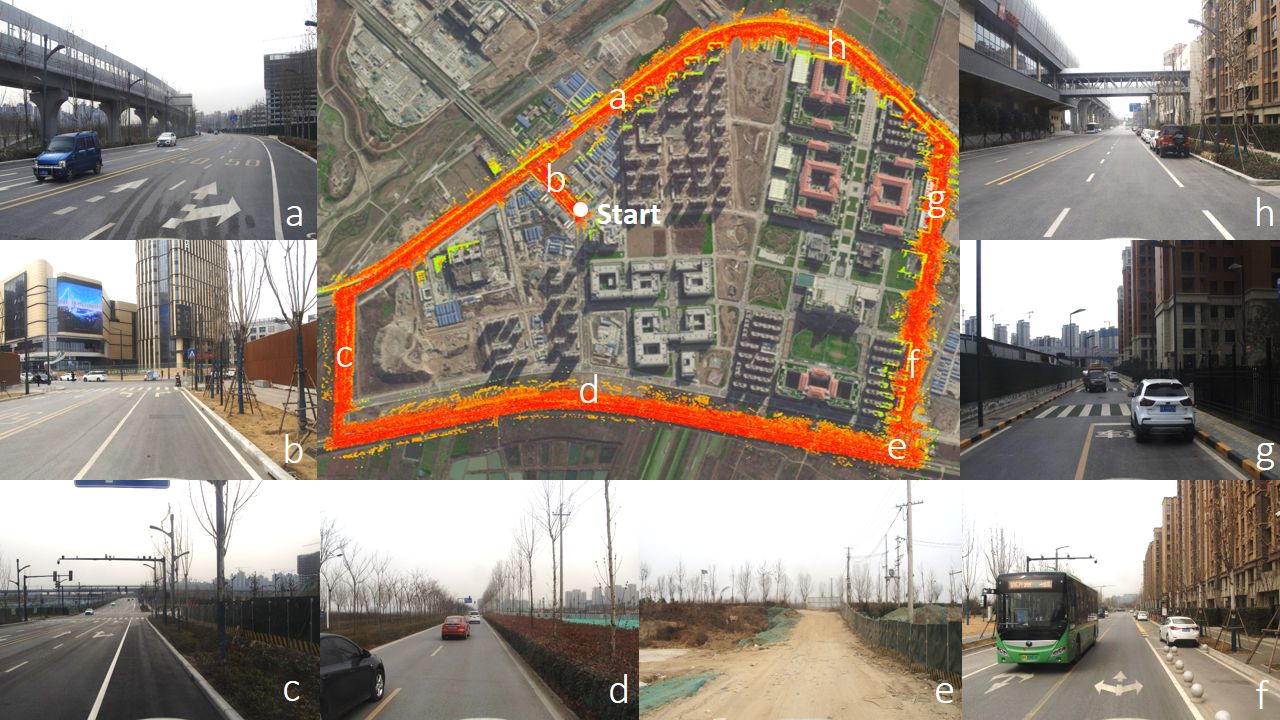}
	\caption{RF-LIO map aligned with satellite image, and some snapshots of the environment.}
	\label{fig: satellite}
\end{figure}

In this experiment, we use three self-collected datasets with only a few moving objects. Since LOAM and LIO-SAM are designed in a static environment, we compare with them to show the performance of RF-LIO in general cases. The urban dataset includes a wide variety of urban terrains: residential area, overpass, construction area, etc. The details and final point cloud map from RF-LIO are shown in Fig. \ref{fig: satellite}. For an intuitive display, the map of RF-LIO is overlaid on a satellite image. The campus dataset is collected from the XJTU campus with multiple pedestrians. As shown in Fig. \ref{fig:test:enviromnet}, the suburban dataset contains various moving vehicles.

To show the benefits of removing moving points before scan-matching, we divide RF-LIO into three different types. When removing moving points first and scan-matching later, our method is referred to as RF-LIO (First). Similarly, when the moving points are removed after scan-matching, our method is referred to as RF-LIO (After). When first removing moving points, then scan-matching, and finally removing moving points again, our method is referred to as RF-LIO (FA). The absolute trajectory error of all methods is shown in Table \ref{table: low_dynamic}. LOAM does not perform well on all three datasets. Therefore here we only compare with LIO-SAM. From the results, it can be seen that RF-LIO (After) and LIO-SAM have similar performance, while RF-LIO (First) and RF-LIO (FA), using removal-first, have a significant improvement. Compared with LIO-SAM, RF-LIO (FA) improves the absolute trajectory accuracy by 36.7\%, 3\%, and 18.3\%, respectively.

\begin{table}[htbp]
	\newcommand{\tabincell}[2]{\begin{tabular}{@{}#1@{}}#2\end{tabular}}
	\centering 
	\caption{Absolute Trajectory RMSE [M] for All Methods Using Low and Medium Dynamic Datasets}
	\label{table: low_dynamic}  
	\tabcolsep 0.05in
	\begin{tabular}{cccccc}
		\hline
		Dataset & LOAM & LIO-SAM & \tabincell{c}{RF-LIO\\(After)} &  \tabincell{c}{RF-LIO\\(First)} & \tabincell{c}{RF-LIO\\(FA)} \\
		\hline
		Urban & 244.19& 10.21 & 10.79 & 7.72 & $\mathbf{6.46}$\\
		Campus & 118.49& 0.66 & 0.66 & 0.64 & $\mathbf{0.61}$ \\
		Suburban & Fail& 1.53 & 1.51& 1.42 &$\mathbf{1.25}$ \\
		\hline
	\end{tabular}
\end{table}

\subsection{Results on High Dynamic Datasets}

\begin{figure}[tbph]
	\centering
	\includegraphics[width=1.0\linewidth,height=0.4\linewidth]{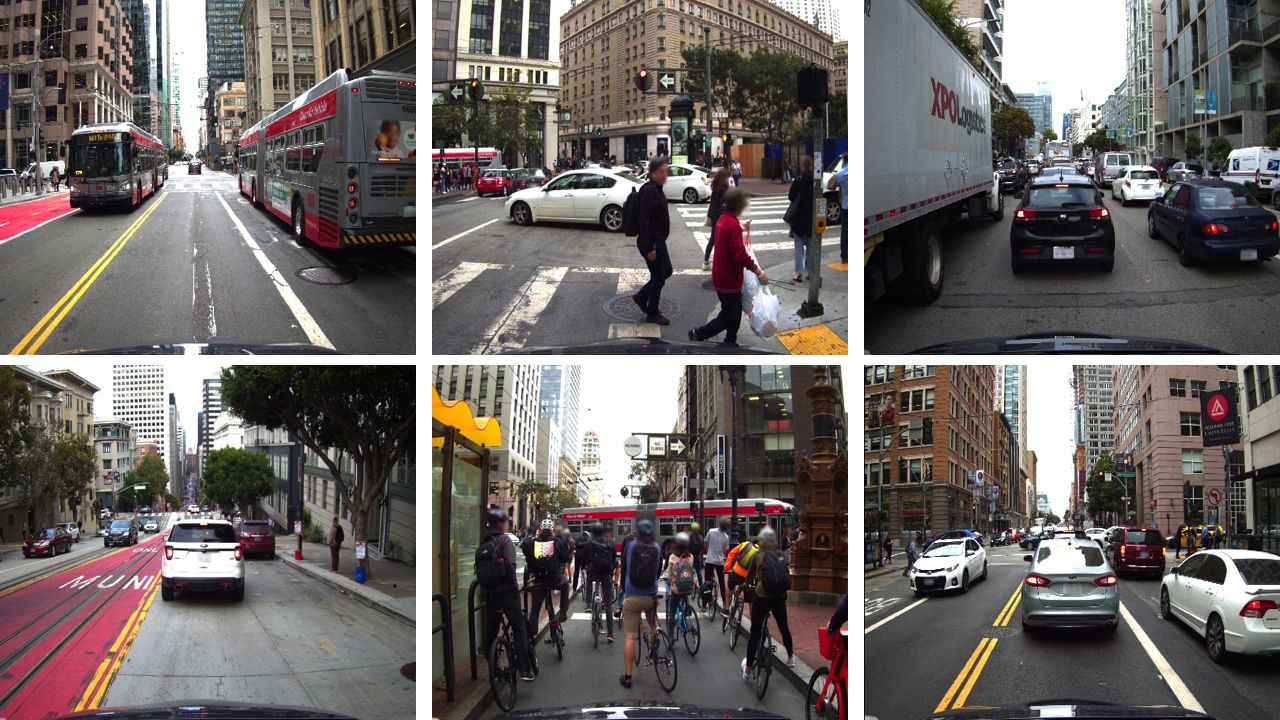}
	\caption{ Some screenshots of the urbanloco datasets. There are many moving objects, which block lidar's FOV and introduce outliers.}
	\label{fig: urbanloco}
\end{figure}

\begin{figure*}[htbp]
	\centering
	\subfigure[Ground truth]{
		\begin{minipage}[t]{0.23\linewidth}
			\centering
			\includegraphics[width=1.0\linewidth,height=0.8\linewidth]{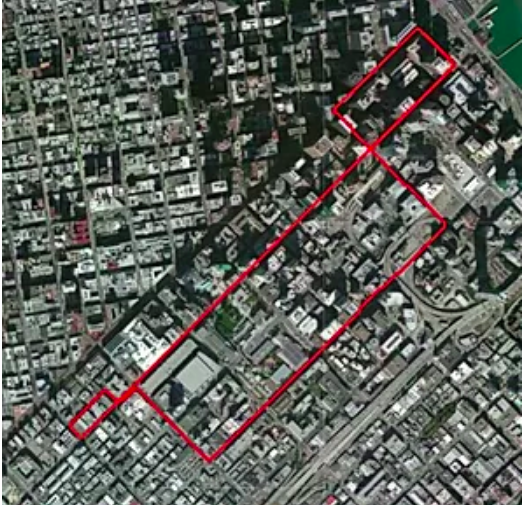}
			\label{Ground truth}
		\end{minipage}%
	}%
	\subfigure[LOAM]{
		\begin{minipage}[t]{0.23\linewidth}
			\centering
			\includegraphics[width=1.0\linewidth,height=0.8\linewidth]{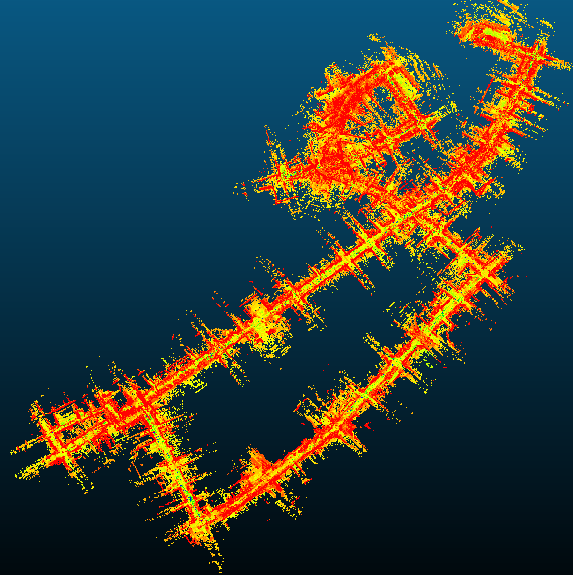}
			\label{loam}
		\end{minipage}%
	}%
	\subfigure[LIO-SAM]{
		\begin{minipage}[t]{0.23\linewidth}
			\centering
			\includegraphics[width=1.0\linewidth,height=0.8\linewidth]{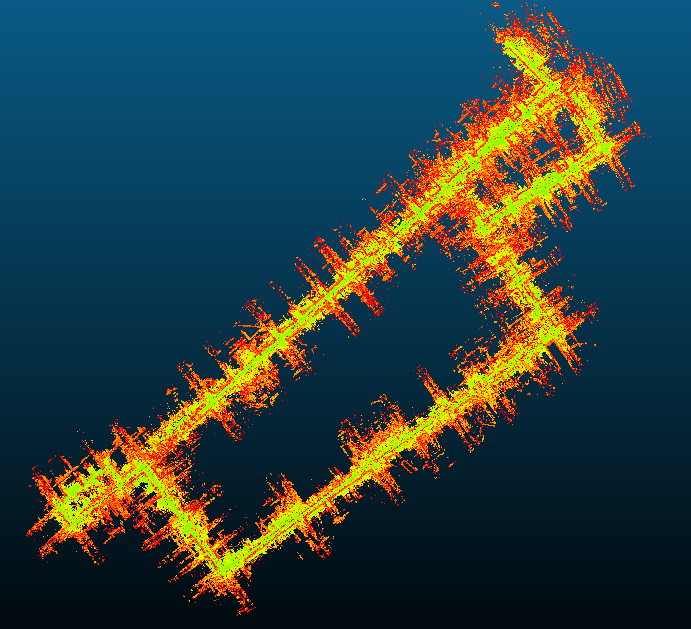}
			\label{lio_sam}
		\end{minipage}
	}%
	\subfigure[RF-LIO]{
		\begin{minipage}[t]{0.23\linewidth}
			\centering
			\includegraphics[width=1.0\linewidth,height=0.8\linewidth]{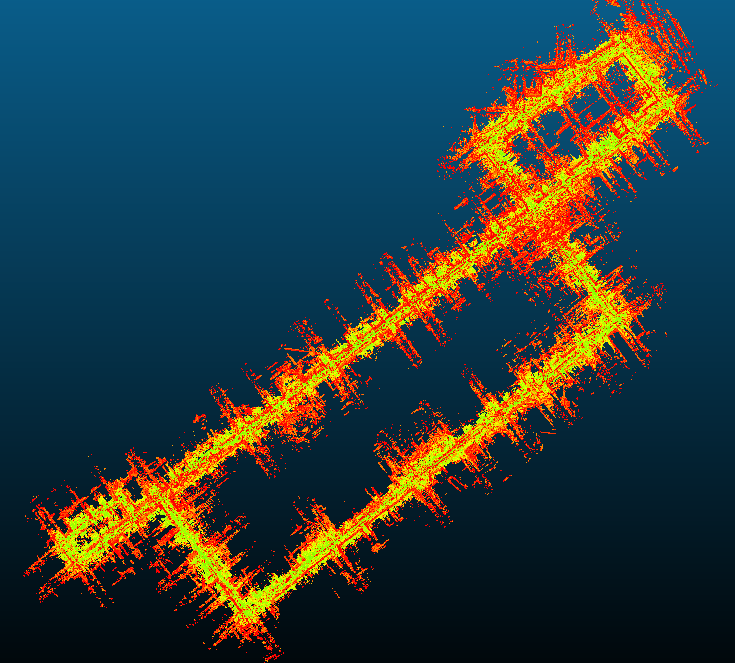}
			\label{RF-LIO}
		\end{minipage}
	}%
	\centering
	\caption{ Mapping results of LOAM, LIO-SAM, and RF-LIO using the CAMarketStreet dataset. }
\end{figure*}

This experiment is designed to demonstrate the superiority of our method in a challenging high dynamic environment. We select the open UrbanLoco datasets as the test dataset, which is collected in highly urbanized scenes with numerous moving objects. Some screenshots of the test environment are shown in Fig. \ref{fig: urbanloco}. In this experiment, when the feature points fall on the moving object, the LOAM map shown in Fig. \ref{loam} diverges in multiple locations. LIO-SAM outperforms LOAM in this test, and its map is shown in Fig. \ref{lio_sam}. Compared with LIO-SAM, RF-LIO has smaller drift and better closed-loop detection performance due to the removal of moving objects. The absolute trajectory error of all methods is shown in Table \ref{table: high_dynamic}.

\begin{table}[htbp]
	\newcommand{\tabincell}[2]{\begin{tabular}{@{}#1@{}}#2\end{tabular}}
	\centering 
	\caption{Absolute Trajectory RMSE [M] for All Methods Using High Dynamic Datasets}
	\label{table: high_dynamic}  
	\tabcolsep 0.05in
	\begin{tabular}{cccccc}
		\hline
		Dataset & LOAM & LIO-SAM & \tabincell{c}{RF-LIO\\(After)} &  \tabincell{c}{RF-LIO\\(First)} & \tabincell{c}{RF-LIO\\(FA)} \\
		\hline
		CAMarktStreet & 203.78& 62.43 &23.98 & $\mathbf{15.83}$& 15.89 \\
		CARussianHill & 175.69& 36.98 & 12.91& 12.44 & $\mathbf{12.17}$\\
		\hline
	\end{tabular}
\end{table}

In the high dynamic datasets, RF-LIO has a more prominent effect on SLAM performance improvement. Compared with LIO-SAM, RF-LIO (After) improves the absolute trajectory accuracy by 61.6\% and 65.1\%, respectively. The experimental results demonstrate that our moving objects removal method can effectively remove dynamic feature points and improve SLAM performance. On these two datasets, RF-LIO (First) outperforms RF-LIO (After) by 34.0\% and 3.6\%, respectively. The results are consistent with the results on low and medium dynamic datasets, but more significant. So we can conclude that removal-first is another effective method to improve SLAM performance in dynamic environments. Because removal-first has removed most of the dynamic points, the trajectory error of RF-LIO (First) and RF-LIO (FA) are not significantly different. In summary, RF-LIO (FA) achieves the best results among all methods.

In practical application, real-time performance is another crucial indicator to evaluate SLAM systems. We test the runtime of RF-LIO (After), RF-LIO (First), and RF-LIO (FA) on all five datasets. The results are shown in Table \ref{table: Runtime}. We see that the runtime of RF-LIO (FA) is less than 100 ms in low dynamic environments, while in high dynamic environments, it is also less than 121 ms. In addition, we note that the runtime of RF-LIO (First) is significantly less than RF-LIO (After), and even the runtime of RF-LIO (FA) is also less than RF-LIO (After). The experimental results show that removal-first takes extra time to remove moving objects, but a clean point cloud can reduce scan-matching time. Although removal-after also removes the moving points, the scan-matching has been completed at this time. Therefore the effect is not as good as removal-first.

\begin{table}[htbp]
	\newcommand{\tabincell}[2]{\begin{tabular}{@{}#1@{}}#2\end{tabular}}
	\centering 
	\caption{ Runtime of RF-LIO for Processing One Scan}
	\label{table: Runtime}  
	\tabcolsep 0.05in
	\begin{tabular}{cccc}
		\hline
		Dataset  & \tabincell{c}{RF-LIO\\(After)} &  \tabincell{c}{RF-LIO\\(First)} & \tabincell{c}{RF-LIO\\(FA)} \\
		\hline
		Urban  & 86 ms & 55 ms & 61 ms\\
		Campus  & 97 ms & 68 ms & 74 ms \\
		Suburban  & 118 ms& 97 ms & 112 ms \\
		CAMarktStreet  &105 ms & 93 ms& 96 ms \\
		CARussianHill  & 134 ms& 107 ms & 121 ms\\
		\hline
	\end{tabular}
\end{table}

\section{CONCLUSIONS}

We propose RF-LIO to perform real-time and robust state estimation and mapping in high dynamic environments. RF-LIO employs moving objects removal-first algorithm combined with tightly-coupled LIO to solve the chicken-and-egg problem of first removing the dynamic points or first scan-matching in high dynamic environments. The proposed adaptive range image moving points removal algorithm does not rely on any prior training data, nor is it limited by the category and number of moving objects. Hence RF-LIO can be applied to various scenes robustly.

However, RF-LIO still has some ongoing works. In a very open environment, if there are no corresponding far points in the surrounding environment, the visibility-based range image method can not remove moving points. Another problem is that when moving objects completely block the FOV of our sensor, the method is not suitable.




%

%

\bibliography{IEEEtranBST/IEEEabrv}

\end{document}